\documentclass[lettersize,journal]{IEEEtran}
\usepackage{amsmath,amsfonts}
\usepackage{algorithmic}
\usepackage{array}
\usepackage[caption=false,font=normalsize,labelfont=sf,textfont=sf]{subfig}
\usepackage{textcomp}
\usepackage{stfloats}
\usepackage{url}
\usepackage{verbatim}
\usepackage{graphicx}

\def\BibTeX{{\rm B\kern-.05em{\sc i\kern-.025em b}\kern-.08em
    T\kern-.1667em\lower.7ex\hbox{E}\kern-.125emX}}
\usepackage{balance}

\usepackage{newfloat}
\usepackage{hhline}
\usepackage{listings}
\usepackage{tikz}
\usepackage{latexsym}
\usepackage[T1]{fontenc}
\usepackage[utf8]{inputenc}
\usepackage{microtype}
\usepackage{inconsolata}
\usepackage{graphicx}
\usepackage{multirow}
\usepackage{algorithmic}
\usepackage{makecell}
\usepackage{algorithm}
\usepackage{amsmath,amsfonts}
\usepackage{textcomp}
\usepackage{url}
\usepackage{array}
\usepackage{amssymb}
\usepackage{booktabs}
\usepackage{xcolor}  
\usepackage{color}
\usepackage{colortbl}
\definecolor{ljcolor}{RGB}{34,139,34}
\usepackage{pifont}
\definecolor{backgroud}{HTML}{DAE3F5}
\definecolor{hdash}{RGB}{191, 191, 191}
\definecolor{unified}{HTML}{4874CB}
\definecolor{attr}{HTML}{EE822F}
\definecolor{graph}{HTML}{E54C5E}
\definecolor{feature}{HTML}{75BD42}
\definecolor{backgroud}{HTML}{DAE3F5}
\definecolor{hdash}{RGB}{191, 191, 191}
\definecolor{ent1}{RGB}{155,180,227}
\definecolor{ent2}{RGB}{239,148,158}
\definecolor{ent3}{RGB}{125,223,215}
\definecolor{ent4}{RGB}{254,217,97}
\definecolor{ent5}{RGB}{169,131,198}
\definecolor{ent6}{RGB}{172,215,142}
\definecolor{ent7}{RGB}{37,227,255}
\definecolor{ent8}{RGB}{0,0,255}
\definecolor{ent9}{RGB}{191,191,191}
\definecolor{ent10}{RGB}{181,139,1}
\definecolor{ent11}{RGB}{0,255,0}
\definecolor{ent12}{RGB}{36,144,135}
\definecolor{ent13}{RGB}{200,29,49}
\definecolor{ent14}{RGB}{249,43,237}
\definecolor{ent15}{RGB}{255,0,0}
\definecolor{uc}{RGB}{77, 191, 249}
\definecolor{bc}{RGB}{5,178,83}
\definecolor{hdash}{RGB}{191, 191, 191}
\definecolor{nl}{HTML}{CBE5B3}
\definecolor{code}{HTML}{F4CCAB}
\usepackage{hyperref}
\hypersetup{
    colorlinks=true,
    citecolor=blue,
    linkcolor=blue, 
    urlcolor=blue 
}

\usepackage{arydshln}
\usepackage[most]{tcolorbox}
\usepackage{bibentry}

\usepackage{CJKutf8}

\begin{document}
\title{Code-MIE: A Code-style Model for Multimodal Information Extraction with Scene Graph and Entity Attribute Knowledge Enhancement}
\author{Jiang Liu, Ge Qiu, Hao Fei, Dongdong Xie, Jinbo Li, Fei Li, Chong Teng, Donghong Ji
\thanks{
Jiang Liu, Ge Qiu, Fei Li, Chong Teng, Donghong Ji are with the Key Laboratory of Aerospace Information Security and Trusted Computing, Ministry of Education, School of Cyber Science and Engineering, Wuhan University (e-mail: liujiang@whu.edu.cn; 2024202210093@whu.edu.cn; lifei\_csnlp@whu.edu.cn; tengchong@whu.edu.cn; dhji@whu.edu.cn).

Hao Fei is with University of Oxford (e-mail: haofei7419@gmail.com).

Dongdong Xie is with the Wuhan Second Ship Design and Research Institute (e-mail: xie.dongdong@163.com).

Jinbo Li is with the China United Network Communications Co., Ltd. Research Institute (e-mail: jinbo@alberta.ca).

{\it{Equal contribution: Jiang Liu, Ge Qiu}}

{\it{Corresponding author: Donghong Ji, Fei Li}}}}

\markboth{Journal of \LaTeX\ Class Files,~Vol.~18, No.~9, September~2020}%
{How to Use the IEEEtran \LaTeX \ Templates}

\maketitle

\begin{abstract}
With the rapid development of large language models (LLMs), more and more researchers have paid attention to information extraction based on LLMs. 
However, there are still some spaces to improve in the existing related methods.
First, existing multimodal information extraction (MIE) methods usually employ natural language templates as the input and output of LLMs, which mismatch with the characteristics of information tasks that mostly include structured information such as entities and relations.
Second, although a few methods have adopted structured and more IE-friendly code-style templates, they just explored their methods on text-only IE rather than multimodal IE. 
Moreover, their methods are more complex in design, requiring separate templates to be designed for each task.
In this paper, we propose a Code-style Multimodal Information Extraction framework (Code-MIE) which formalizes MIE as unified code understanding and generation. 
Code-MIE has the following novel designs:
(1) Entity attributes such as gender, affiliation are extracted from the text to guide the model to understand the context and role of entities.
(2) Images are converted into scene graphs and visual features to incorporate rich visual information into the model.
(3) The input template is constructed as a Python function, where entity attributes, scene graphs and raw text compose of the function parameters.
In contrast, the output template is formalized as Python dictionaries containing all extraction results such as entities, relations, etc.
To evaluate Code-MIE, we conducted extensive experiments on the M$^3$D, Twitter-15, Twitter-17, and MNRE datasets. The results show that our method achieves state-of-the-art performance compared to six competing baseline models, with 61.03\% and 60.49\% on the English and Chinese datasets of M$^3$D, and 76.04\%, 88.07\%, and 73.94\% on the other three datasets.
\end{abstract}


\section{Introduction}
\label{Introduction}
With the rapid development of the Internet, in addition to the emergence of text information, other modal information such as images, videos and audio has appeared in large quantities. These modal information have been demonstrated by many studies to promote IE tasks \cite{multimodal_ie_1,multimodal_ie_2,multimodal_ie_3}.
Using large language model (LLM) to complete information extraction tasks is receiving increasing attention, 
they mostly used natural language templates to express these structured tasks \cite{uie_more_study_1,uie_more_study_2}. 
However, such approach suffers some potential shortages:
(1) The output of information extraction tasks is often structured and difficult to convert into plain text \cite{CodeIE}.
(2) Even when these structured information are represented using specially designed formal languages, these languages are difficult for large language models to understand and follow \cite{code_uie_1}, leading to more hallucination (cf. Section \ref{Hallucination Error}).
Therefore, researchers began to use templates with code-style to express structured information \cite{code_uie_1,code_uie_2}. Code itself is a structured representation, which can be naturally used to express structured tasks such as IE. In addition, code has a complete logical structure, so it is easier for LLMs to understand and follow than manually-designed templates. 

\begin{figure}[!t]
\centering
\includegraphics[width=0.35\textwidth]{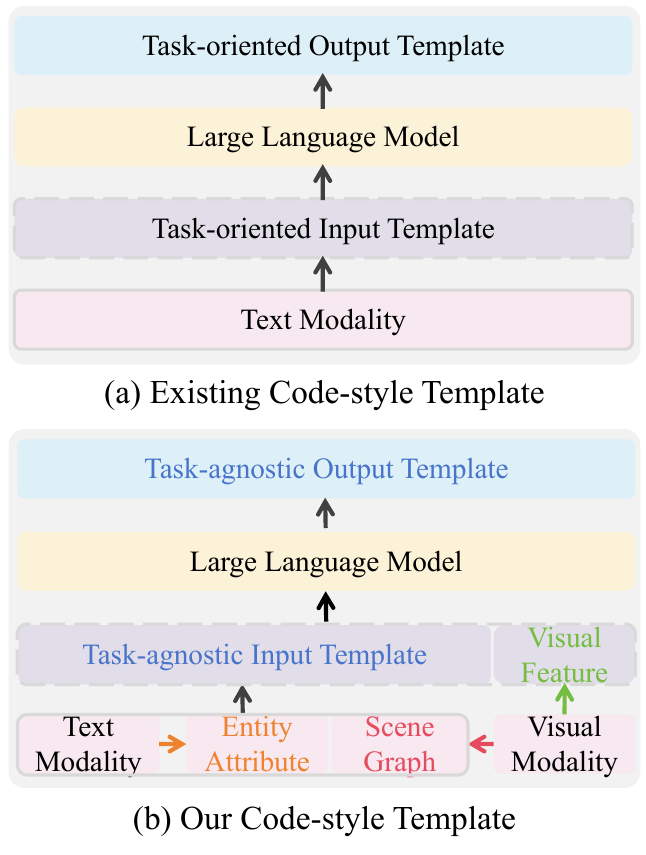}
\caption{
The difference between our code-style template and existing code-style templates.
}
\label{fig1}
\end{figure}

Although code-style templates are more suitable for representing structured information, recent studies have not explored them in multimodal scenarios. 
Therefore, this paper designs a more concise, task-agnostic Code-style model for Multimodal Information Extraction, named Code-MIE, which is used to perform multimodal information extraction tasks in a unified method. 
As shown in Figure \ref{fig1}, our code-style template differs from existing code-style templates \cite{CodeIE,code_uie_1} in three ways: 
(1) Our approach extends code-style templates to the multimodal domain, integrating visual information from implicit and explicit perspectives through \textcolor{feature}{\textbf{visual features}} and \textcolor{graph}{\textbf{scene graphs}}; 
(2) Our template incorporates \textcolor{attr}{\textbf{entity attribute}}, thereby enhancing the reasoning ability and especially facilitating relation extraction tasks requiring deep reasoning; 
(3) Our code-style template is \textcolor{unified}{\textbf{task-agnostic}}, which means we do not need to design a separate template for each task. This design is not only concise but also makes better use of the relations between different tasks.

The illustration of our method is shown in Figure \ref{frame}. Specifically, we first utilize Qwen3-Max \cite{qwen3} to extract entity attributes from the text. Then we utilize Qwen3-VL-235B \cite{qwen3} and ViT \cite{vit} to convert image information into scene graphs and visual features. 
Then, we defined a Python function that takes raw text, entity attributes, and scene graphs as input to construct an input template. The input template is encoded into text features and then concatenated with visual features as input to the LLM. Finally, the LLM outputs the results of all tasks in code format.

We conducted experiments on four datasets, as follows:
M$^3$D \cite{m3d} is a video-based multimodal, multilingual, and multitasking information extraction dataset.
Twitter-15 \cite{tweet-2015} and Twitter-17 \cite{tweet-2017} are image-based multimodal named entity recognition datasets.
MNRE \cite{mre} is an image-based multimodal relation extraction dataset.
Our Code-MIE framework outperforms all baseline models of the M$^3$D dataset in both the English and Chinese settings, achieving the F1s of 61.03\% and 60.49\%, respectively. Similarly, Code-MIE outperforms all baseline models on Twitter-15, Twitter-17 and MNRE datasets, achieving the F1s of 76.04\%, 88.07\% and 73.94\%.
In summary, our contributions are highlighted as below:

$\bullet$ We designed a more concise, task-agnostic Code-style model for Multimodal Information Extraction, named Code-MIE, which completes multiple MIE tasks sequentially, including entity recognition, entity chain extraction, relation extraction, and visual localization, thereby effectively utilizing the relations between tasks and reducing formatting errors and hallucination errors.

$\bullet$ We explicitly and implicitly integrate visual knowledge (e.g., objects and their relations in images) into our framework using scene graph knowledge and visual features. This method not only extends the code-style template from the plain text domain to the multimodal domain, but also further improves the performance of MIE tasks. Additionally, We defined at least two entity attributes for each entity type and incorporated this knowledge into our framework to enhance the reasoning capabilities of model.

$\bullet$ Experiments show that our method significantly outperforms six baseline models on four public datasets, achieving a performance improvement of up to 6.94\% compared to natural language style templates and up to 5.49\% compared to existing code-style templates.

\section{Related Work}
\label{Related Work}
\subsection{Multimodal Information Extraction} 

Lu et al. \cite{tweet-2015} and Zhang et al. \cite{tweet-2017} proposed the MNER dataset. Later, Zheng et al. \cite{mre} proposed the MRE dataset. Research on MIE mainly falls into two directions: one focuses on the alignment between different modalities. Yu et al. \cite{related_work_ner_1} first focused on the alignment problem of images and texts. Later, most researchers \cite{related_work_ner_entity_level,related_Information,multimodal_ie_2} achieve modal alignment by constructing heterogeneous graphs. In addition, Liu et al. \cite{Hierarchical_mner} proposed a hierarchical framework for dynamically aligning images and text, while Zheng et al. \cite{related_Translation} and Jia et al. \cite{mrc_mner} viewed modal alignment as a translation task and a machine reading comprehension task. Other research directions enhance model performance by introducing external knowledge. For example, some researchers introduce additional knowledge by generating additional images \cite{Mixup_Image_Augmentation_mner} or entity information \cite{Entity_Attributes_ent_rel,retrieval_Classification}. However, Wang et al. \cite{more} used a retrieval method to query external knowledge bases.
Overall, existing MIE methods are mostly based on task-customized architectures, relying on explicit cross-modal alignment or external knowledge injection, and typically do not utilize the structured generation capabilities of LLMs. We are the first to introduce code-style templates into multimodal information extraction and effectively integrate information such as scene graphs and entity attributes.

\subsection{Code-style Information Extraction} 
Li et al. \cite{CodeIE} first proposed to rewrite the structured output in the form of code rather than natural language. Guo et al. \cite{code_uie_2} proposed a code generation framework Code4UIE. It designed a unified code-style pattern for various IE tasks through Python classes and adopted a retrieval enhancement mechanism to fully utilize the in-context learning capability of LLM. Bi et al. \cite{Codekgc} developed pattern-aware cues that effectively exploited the semantic structure in knowledge graphs and adopted a generative approach to enhance the performance. Li et al. \cite{code_uie_1} proposed KnowCoder, a code-style pattern library covering various knowledge types, and improved its pattern understanding ability through pre-training. Later, Zuo et al. \cite{Knowcoder-x} further enhanced multilingual and cross-language capabilities of KnowCoder.
While code-style templates have proven superior to natural language templates in plain text information extraction, their application has not yet extended to the multimodal domain, and their templates are task-dependent. Our Code-MIE framework is the first to extend code-style templates to the multimodal domain, integrating visual features, scene graphs, and entity attributes through a task-agnostic Python function template, thereby achieving joint modeling and performance improvement for multiple MIE tasks.

\begin{figure*}[!t]
\centering
\includegraphics[width=1\textwidth]{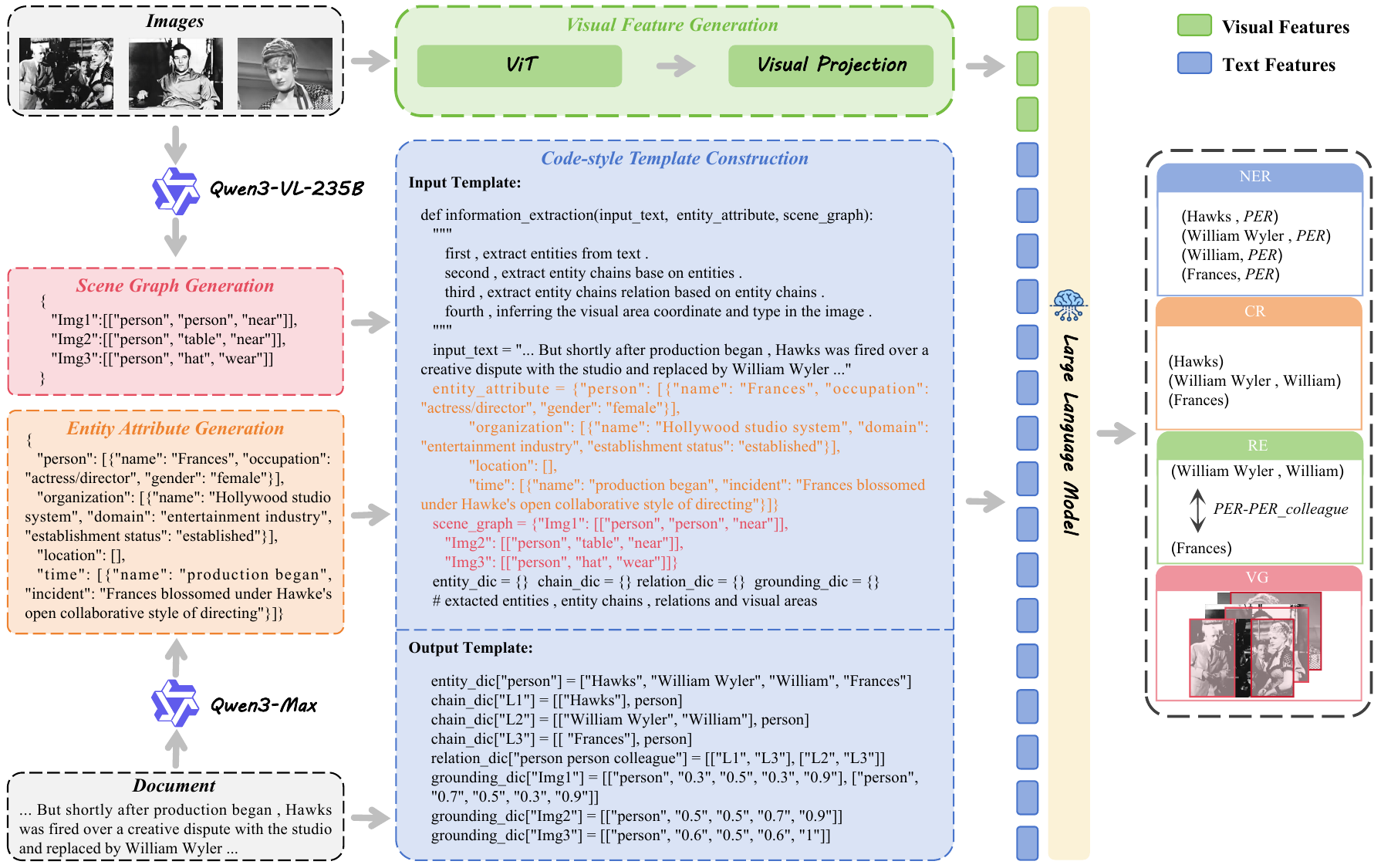}
\caption{
The framework of Code-MIE.
} 
\label{frame}
\end{figure*}

\section{Methodology}
\label{Methodology}
The illustration of our method is shown in Figure \ref{frame}. It mainly consists of five steps: entity attribute generation, scene graph generation, visual feature generation, code-style template construction, and fine-tuning.

\begin{table}[!t]
\fontsize{7}{9}\selectfont
\setlength{\tabcolsep}{0.8mm}
\centering
\caption{Detailed set of entity attributes.
}
\resizebox{0.3\textwidth}{!}{
\begin{tabular}{l c l}
\hline
\bf PER & \phantom{} & \makecell[l]{name, occupation, gender, \\nationality, marital status,\\ place of birth, place of death}\\
\hline
\bf LOC & \phantom{} & \makecell[l]{name, type, function}\\
\hline
\bf ORG & \phantom{} & \makecell[l]{name, type, establishment\\ status, affiliation, domain}\\
\hline
\bf TIME & \phantom{} & \makecell[l]{name, incident}\\
\hline
\end{tabular}
}
\label{tab:attr_set}
\end{table}

\subsection{Entity Attribute Generation}
\label{Entity Attribute Generation}
We use Qwen3-Max \cite{qwen3} to extract entity attributes. It supports both English and Chinese, and is currently the most advanced text-to-text language model. Given a text $X$ and a set $T = \{t_{1}, t_{2},\dots, t_{n}\}$ containing $n$ entity types. For the $i$-th entity type ${t_{i}}$, we define its attribute set $A_{t_{i}} = \{a_{i,1}, a_{i,2},\dots, a_{i,m_{i}}\}$, where $m_{i}$ represents the number of attributes. Table \ref{tab:attr_set} shows the detailed set of entity attributes for each entity type\footnote{The entity types in datasets Twitter-15 and Twitter-17 include PER, LOC, ORG, and OTHER/MISC. Since OTHER/MISC contains many different types and it is difficult to define its attributes uniformly, we have not defined any attributes for this type.}. We will explain the meaning of each attribute in detail. 
\begin{itemize}
    \item \textit{name:} The specific names of person, location, organization, and time must originate from the entity being labeled.
    \item \textit{occupation:} represents a person's profession, such as editor, businessman, singer, musician, actor, entertainer, and actress.
    \item \textit{gender:} represents a person's gender, including male and female.
    \item \textit{nationality:} represents a person's nationality, such as American, British, Russian, German, French, and Irish.
    \item \textit{marital status:} represents a person's marital status, and its value is derived from a summary of the original text.
    \item \textit{place of birth:} represents a person's place of birth, usually a country or more specifically a province/state, etc.
    \item \textit{place of death:} represents a person's place of death, which is the same as their place of birth, usually a country or more specifically a province/state, etc.
    \item \textit{type:} represents the type of location or organization, location types include state, city, country, region, continent, street, etc.; organization types include high school, military unit, film studio, military academy, research institution, private enterprise, company, religious organization, etc.
    \item \textit{function:} represents the function or role of a location, and its value is derived from a summary of the original text.
    \item \textit{domain:} represents the domain which an organization belongs, including education, media, military, entertainment, music, defense, record industry, theater, religion, literature/arts, etc.
    \item \textit{establishment status:} represents the current status of an organization, including establish and disintegrate.
    \item \textit{affiliation:} represents the country or organization to which an organization is affiliated, including the United States, Germany, Russia, the United Kingdom, North Carolina State University, the United States Congress, and the Democratic Party, among others.
    \item \textit{incident:} represents the event occurring now, and its value is derived from a summary of the original text.
\end{itemize}

Then, we generate entity attributes based on the following three steps.

\begin{table*}[!t]
\fontsize{7}{9}\selectfont
\setlength{\tabcolsep}{0.8mm}
\centering
\caption{Entity attribute examples.
}
\includegraphics[width=0.9\textwidth]{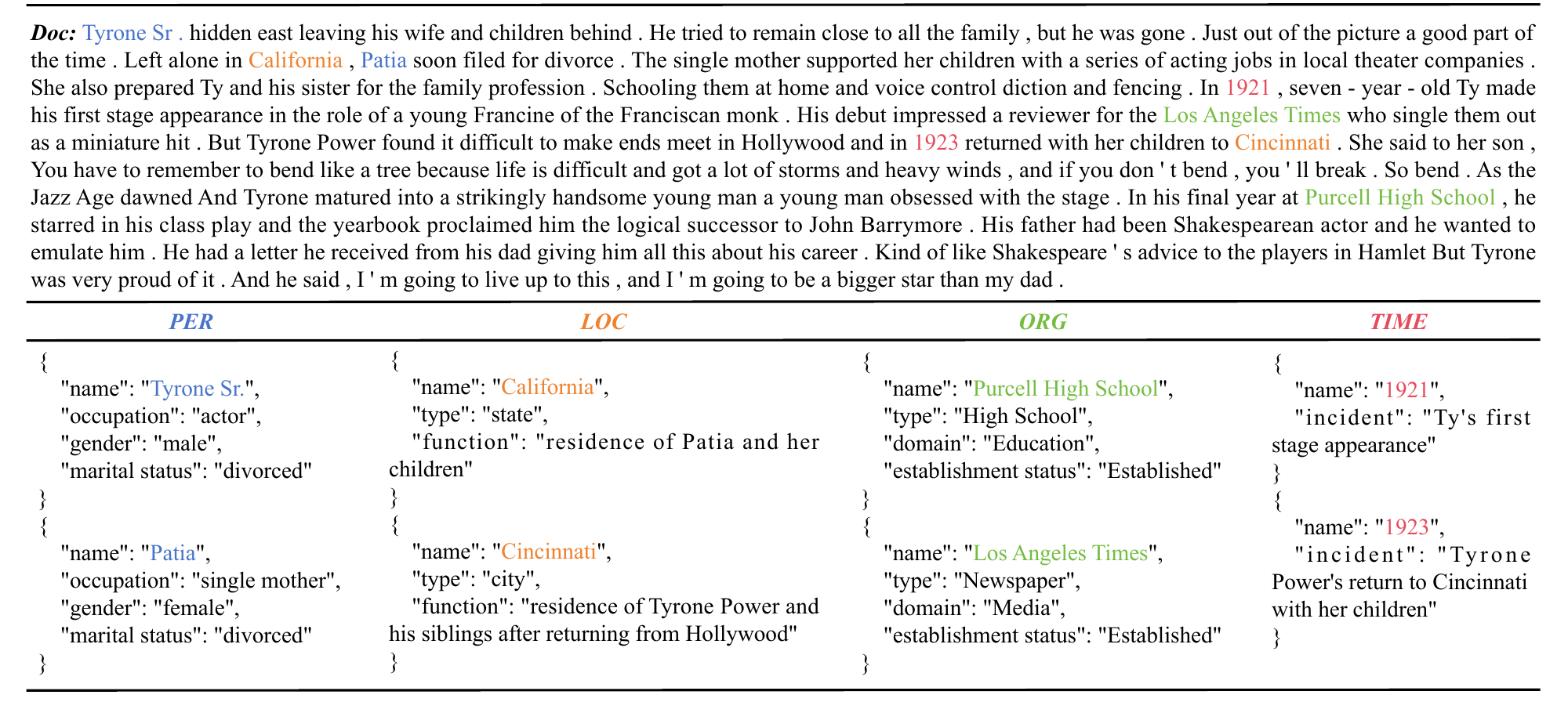}
\label{tab:attr_example_2}
\end{table*}

\textit{\textbf{Step 1:}} For each entity type, we construct a template for generating entity attributes, which takes the following form:
\begin{tcolorbox}[colback=gray!5,colframe=orange!60,breakable]
\quad\quad Based on the given text $X$, analyze the possible attributes of the $t_{i}$ entity type, such as $a_{i,1}$, $a_{i,2}$, \dots, $a_{i,m_{i}}$. If it is not mentioned in the text, display "not mentioned". Note the following:

\quad\quad1. Output all results in the format ($a_{i,1}$, $a_{i,2}$, \dots, $a_{i,m_{i}}$).

\quad\quad2. Output as many results as possible.

\quad\quad3. Do not perform any other additional analysis.
\end{tcolorbox}

\textit{\textbf{Step 2:}} We use Qwen3-Max to generate results for each entity type, as shown below:
\begin{tcolorbox}[colback=gray!5,colframe=orange!60,breakable]
($v_{i,1,1}$, $v_{i,2,1}$, \dots, $v_{i,m_{i},1}$)

($v_{i,1,2}$, $v_{i,2,2}$, \dots, $v_{i,m_{i},1}$)

\dots\dots

($v_{i,1,k_{i}}$, $v_{i,2,k_{i}}$, \dots, $v_{i,m_{i},k_{i}}$)
\end{tcolorbox}
\noindent where $v_{i,j,k}$ represents the $k$-th value of the $j$-th attribute of the $i$-th entity type $t_{i}$, and $k_{i}$ represents the number of attribute values in the $i$-th entity type. We will repeat this step three times, combine the results from the three steps, and remove duplicates. 

\textit{\textbf{Step 3:}} Since we don't restrict the values of each attribute, LLM can freely generate any possible values. Secondly, although the second step is executed multiple times, LLM still produces a small number of hallucination errors. Therefore, we perform simple post-processing and manual correction on the generated results. Our processing rules follow these three points:

1) Remove unreasonable values, such as treating a person name as an organization name or location name, or treating movies/TV shows/literary works as organizations.

2) Only retain the mentioned attributes.

3) The final result must have at least two attributes and must include a "name" field.

Finally, we will obtain structured entity attributes $X^{EA}$:
\begin{tcolorbox}[colback=gray!5,colframe=orange!60,breakable]
$t_{i}$: [\{$a_{i, 1}$: $v_{i,1,1}$, $a_{i, 2}$: $v_{i,2,1}$, \dots, $a_{i, m_{i}}$: $v_{i,m_{i},1}$\}, \dots, \{$a_{i, 1}$: $v_{i,1,k_{i}}$, $a_{i, 2}$: $v_{i,2,k_{i}}$, \dots, $a_{i, m_{i}}$: $v_{i,m_{i},k_{i}}$\}]
\end{tcolorbox}

\noindent Table \ref{tab:attr_example_2} shows an example of generated entity attributes.

\subsection{Scene Graph Generation} 
\label{Scene Graphs Generation}
A scene graph is a structured representation method used to transform information in a visual scene into a graph structure, including objects, their attributes, and the relations between them \cite{Scene_Graph}. Therefore, we generate scene graph knowledge of images and incorporate it into our framework to explicitly capture the visual objects and their relations in the image, which can assist our framework in making decisions. Specifically, for a given image set $G=\{g_{1},g_{2},\dots,g_{q}\}\in\mathbb{R}^{q}$ containing $q$ images, We use Qwen3-VL-235B \cite{qwen3} and generate its scene graph for each image in three steps.

\textit{\textbf{Step 1:}} We construct a template for generating scene graphs, taking one image as input each time, in the following format:
\begin{tcolorbox}[colback=gray!5,colframe=red!60,breakable]
\quad\quad Analyze the scene graph contained in the given image. Note the following:

\quad\quad1. Output all results in the format (subject, object, relation).

\quad\quad2. Output as many results as possible.

\quad\quad3. Do not perform any other additional analysis.
\end{tcolorbox}

\textit{\textbf{Step 2:}} We use Qwen3-VL-235B to generate results for $i$-th image, as shown below:
\begin{tcolorbox}[colback=gray!5,colframe=red!60,breakable]
($s_{i,1}$, $o_{i,1}$, $r_{i,1}$)

($s_{i,2}$, $o_{i,2}$, $r_{i,2}$)

\dots\dots

($s_{i,z_{i}}$, $o_{i,z_{i}}$, $r_{i,z_{i}}$)
\end{tcolorbox}
\noindent where $s_{i,j}$, $o_{i,j}$ and $r_{i,j}$ represent the subject, object and relation of the $j$-th scene graph triple in the $i$-th image, and $z_{i}$ representsthe number of relation triples in the $i$-th image. We will repeat this step three times, combine the results from the three steps, and remove duplicates. 

\textit{\textbf{Step 3:}} For post-processing of scene graphs, our requirements are more flexible. Unlike entity attributes, the subjects and objects in a scene graph can be person, locations, organizations, or even some items (such as jewelry, tables, hats, etc.) or scenery (such as clouds, trees, etc.). The relations are mainly action relations (such as hugging, holding, or wearing) and positional relations (such as nearby, above, left, etc.). Finally we obtain structured scene graphs $X^{SG}$:
\begin{tcolorbox}[colback=gray!5,colframe=red!60,breakable]
Img$_{i}$: [[$s_{i,1}$, $o_{i,1}$, $r_{i,1}$], [$s_{i,2}$, $o_{i,2}$, $r_{i,2}$], \dots, [$s_{i,z_{i}}$, $o_{i,z_{i}}$, $r_{i,z_{i}}$]]
\end{tcolorbox}
\noindent Table \ref{tab:sg_example} shows some examples of generated scene graphs.

\begin{table}[!t]
\fontsize{7}{9}\selectfont
\setlength{\tabcolsep}{0.8mm}
\caption{Scene graph examples.}
\centering
\includegraphics[width=0.43\textwidth]{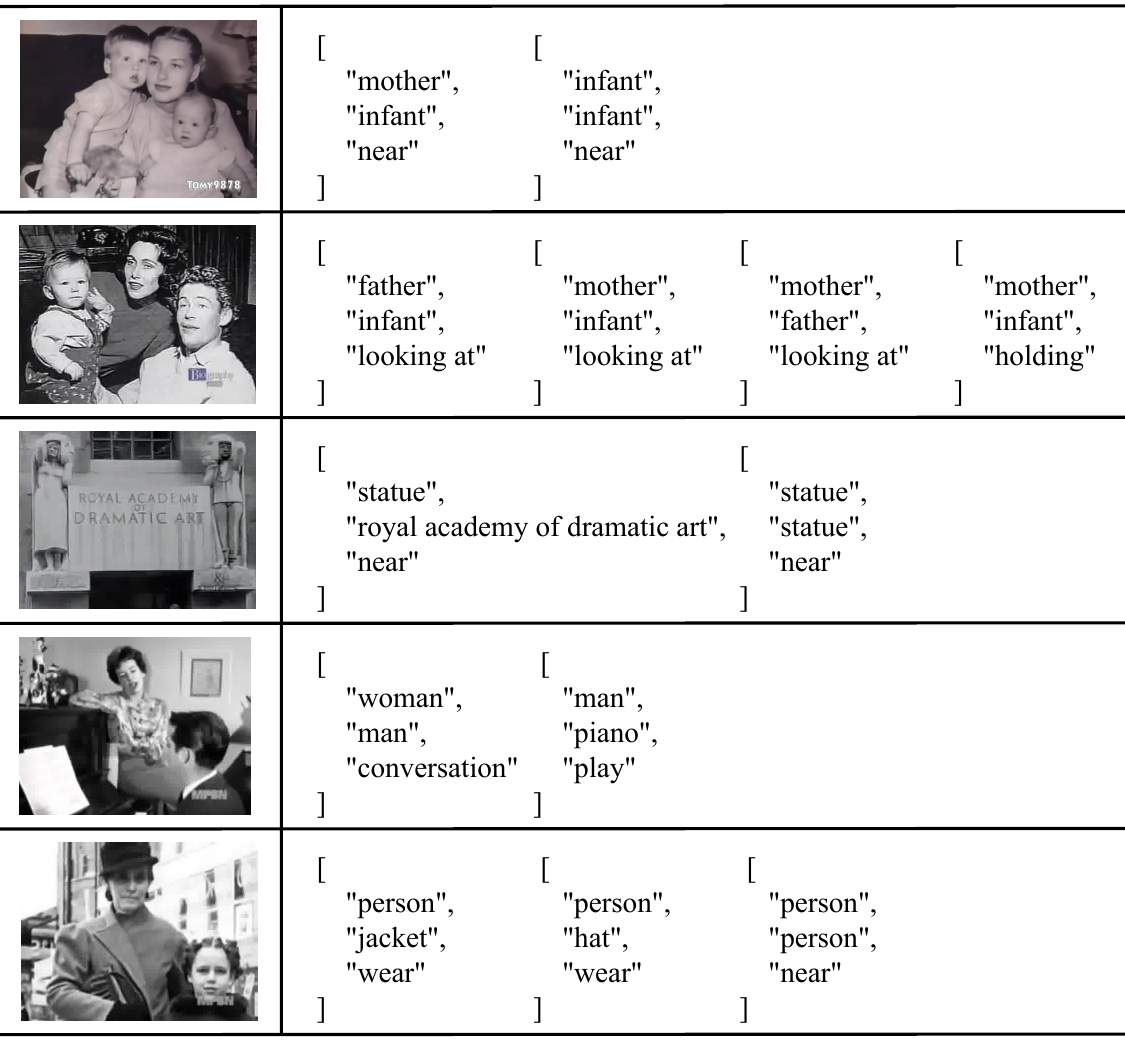}
\label{tab:sg_example}
\end{table}

\subsection{Visual Feature Generation} We use visual pre-training model ViT \cite{vit} to encode the image. ViT first divides each image into $n_{p}$ patches and then encode the linearized patches, so we can get the image embedding $\boldsymbol{H^{g}}$:
\begin{equation}
\label{deqn_ex4a}
\begin{gathered}
\boldsymbol{H^{g}}=\{\boldsymbol{H}^{g}_{1},\boldsymbol{H}^{g}_{2},\dots,\boldsymbol{H}^{g}_{q}\}\in\mathbb{R}^{q \times n_{p} \times d_{g}},\\
\boldsymbol{H}^{g}_{i}=\{\boldsymbol{h}^{g}_{i,1},\boldsymbol{h}^{g}_{i,2},\dots,\boldsymbol{h}^{g}_{i,n_{p}}\}\in\mathbb{R}^{n_{p} \times d_{g}},\\
\end{gathered}
\end{equation}
where $\boldsymbol{h}^{g}_{i,j}\in\mathbb{R}^{d_{}h}$ represents the embedding of the $j$-th patch of the $i$-th image. $d_{g}$ represents the dimension of visual embeddings. Then we fuse the intra-frame features using average pooling (AvgPool). In addition, we introduce positional embeddings $\boldsymbol{H}^{p}$ to account for the
inter-image frame temporal information. Finally, we obtain the visual feature $\boldsymbol{H}^{v}\in\mathbb{R}^{q\times d_{g}}$:
\begin{equation}
\label{deqn_ex4a}
\begin{gathered}
\boldsymbol{H^{v}}=\text{AvgPool}(\boldsymbol{H^{g}})+\boldsymbol{H}^{p}.\\
\end{gathered}
\end{equation}

\subsection{Code-style Template Construction}
Next, we will introduce how to build a code-style template. After getting the entity attribute $X^{EA}$ and the scene graph $X^{SG}$, we can build our code-style template $X^{input}$. We use Python functions to build our input templates. We use \texttt{information\_extraction} as the name of the Python function and add a documentation string to describe the function's function. We assign the input document $X$, the entity attribute $X^{EA}$, and the scene graph $X^{SG}$ to the variables \texttt{input\_text}, \texttt{entity\_attribute} and \texttt{scene\_graph}, respectively, which are the input parameters of the function. Then, we initialize four empty dictionaries, \texttt{entity\_dic}, \texttt{chain\_dic}, \texttt{relation\_dic} and \texttt{grounding\_dic}, which are used to store entities, entity chains, relations, and visual areas, respectively. Our input template can be expressed as:

\begin{tcolorbox}[colback=gray!5,colframe=unified!60,breakable]
def information\_extraction(input\_text, entity\_attribute, scene\_graph):
    
\quad"""first , extract entities from text .
        
\quad\quad second , extract entity chains base on entities .
        
\quad\quad third , extract entity chains relation based on entity chains .
        
\quad\quad fourth , inferring the visual area coordinate and type in the image based on the scene graph ."""
    
\quad input\_text = $X$

\quad entity\_attribute = $X^{EA}$

\quad scene\_graph = $X^{SG}$

\quad entity\_dic = \{\} 

\quad chain\_dic = \{\} 

\quad relation\_dic = \{\} 

\quad grounding\_dic = \{\}
    
\quad \# extacted entities , entity chains , relations and visual areas
\end{tcolorbox}

\noindent For the output template $X^{output}$, use Python elements to build, such as lists, dictionaries, etc. For the entity recognition results, we put $n_{e}$ entities with the same entity type $t^{e}$ into a list and add them to \texttt{entity\_dic}: 
\begin{tcolorbox}[colback=gray!5,colframe=unified!60,breakable]
entity\_dic[$t^{e}$] = [$e_{1}, e_{2},\dots,e_{n_{e}}$]
\end{tcolorbox}

\noindent For the entity chain recognition result, we add the entity chain $l = \{e^{l}_{1}, e^{l}_{2}, \dots, e^{l}_{n_{l}}\}$ with $n_{l}$ entities to \texttt{chain\_dic}:
\begin{tcolorbox}[colback=gray!5,colframe=unified!60,breakable]
chain\_dic[$ID(l)$] = [$l$, $t^{l}$]
\end{tcolorbox}

\noindent where $ID(\cdot)$ represents the ID of the entity chain to simplify the output template of the relation extraction task, and $t^{l}$ represents the type of entity chain. For the relation extraction result, we put $n_{r}$ entity chain pairs with the same relation type $t^{r}$ into a list and add them to \texttt{relation\_dic}:
\begin{tcolorbox}[colback=gray!5,colframe=unified!60,breakable]
relation\_dic[$t^{r}$] = [[$ID(l^{sub}_{1})$, $ID(l^{obj}_{1})$], [$ID(l^{sub}_{2})$, $ID(l^{obj}_{2})$], $\dots$, [$ID(l^{sub}_{n_{r}})$, $ID(l^{obj}_{n_{r}})$]]
\end{tcolorbox}

\noindent where entity chain pair is also a list containing the ID of the subject entity chain and the object entity chain. For the visual grounding result, we put $n_{u}$ visual regions into a list and add them to \texttt{grounding\_dic}:
\begin{tcolorbox}[colback=gray!5,colframe=unified!60,breakable]
grounding\_dic[Img$_{i}$] = [$u_{i,1}$, $u_{i,2}$ $\dots$, $u_{i,n_{u}}$]
\end{tcolorbox}

\noindent where the $j$-th visual region $u_{i,j}=[t^{u}_{i,j},c_{i,j}^{horiz},c_{i,j}^{vert},c_{i,j}^{wd},c_{i,j}^{ht}]$, and $t^{u}_{i,j}$, $c_{i,j}^{horiz}$, $c_{i,j}^{vert}$, $c_{i,j}^{wd}$ and $c_{i,j}^{ht}$ represent the type of visual target, the central horizontal coordinate, central vertical coordinate, width and height of the visual target bounding box respectively.

\begin{table*}[!t]
\fontsize{5}{7}\selectfont
\setlength{\tabcolsep}{0.8mm}
\centering
\caption{Statistics for M$^{3}$D. (Doc.: Document, Sent.: Sentence, Ent.: Entity, Rel.: Relation, Cha.: Entity Chain, Gro.: Visual Grounding)}
\resizebox{0.8\textwidth}{!}{
\begin{tabular}{cccccccccccccc}
\hline
&\multicolumn{6}{c}{\bf EN} &\phantom{}&\multicolumn{6}{c}{\bf ZH}\\
\cline{2-7}\cline{9-14}
&\bf Doc.&\bf Sent.&\bf Ent.&\bf Cha.&\bf Rel.&\bf Gro.&\phantom{}&\bf Doc.&\bf Sent.&\bf Ent.&\bf Cha.&\bf Rel.&\bf Gro.\\
\hline
\bf Train & 1,644 &28,525& 27,843 &15,031&11,410&9,812&\phantom{}&1,629&11,606&22,043&11,481&8,372&3,726\\
\bf Dev & 205 & 3,616&3,457 &1,885&1,388&1,171&\phantom{}&203&1,421&2,685&1,397&985&505\\
\bf Test &207 &3,649&3,587&1,882&1,364&1,198&\phantom{}&205&1,454&2,728&1,419&1,049&482\\
\hline
\end{tabular}
}
\label{tab:dataset_Statistics_m3d}
\end{table*}

\begin{table}[!t]
\fontsize{5}{7}\selectfont
\setlength{\tabcolsep}{0.8mm}
\centering
\caption{Statistics for Twitter-15, Twitter-17 and MNRE.}
\resizebox{0.48\textwidth}{!}{
\begin{tabular}{c c c c c c c c c c c}
\hline
&\multicolumn{2}{c}{\bf Twitter-15} &\phantom{}&\multicolumn{2}{c}{\bf Twitter-17}& \phantom{}&\multicolumn{3}{c}{\bf MNRE}\\
\cline{2-3}\cline{5-6}\cline{8-10}
&\bf Sent.&\bf Ent.&\phantom{}&\bf Sent.&\bf Ent.&\phantom{}&\bf Sent.&\bf Ent.&\bf Rel.\\
\hline
\bf Train & 4,000 &6,123& \phantom{}&3,373&6,003&\phantom{}&7,330&2,4494&12,247\\
\bf Dev & 1,000 & 1,524&\phantom{}&723&1,313&\phantom{}&931&3,248&1,624\\
\bf Test &3,257 &5,085&\phantom{}&723&1,339&\phantom{}&914&3,228&1,614\\
\hline
\end{tabular}
}
\label{tab:dataset_Statistics_mnre}
\end{table}

\subsection{Fine-tuning}
Our framework has two goals in the fine-tuning stage, one is  to train LLMs with information extraction capability, and the other is to train LLMs with multimodal alignment capability. 

\subsubsection{Information Extraction Capability} 
To make our framework IE-capable, we fine-tune the backbone LLM using the code-style template that we have built. Since fine-tuning LLM directly will consume a lot of time and space, we use LoRA \cite{LoRA} to fine-tune the model by only fine-tuning a small subset of parameters without changing the overall LLM architecture.

\subsubsection{Multimodal Alignment Capability} 
In addition, our framework also needs to achieve multimodal information alignment. Specifically, we first encode the template using the encoding layer of the LLM to obtain text features. Then, we adjust the dimensions of the visual features through the visual projection. Finally, we concatenate the visual features and text features and input them into the LLM. During fine-tuning, the ViT \cite{vit} is frozen, and only the visual projection and the backbone LLM are fine-tuned.

\section{Experiment Setups}
\label{Experiment Setups}
\subsection{Datasets} 
We tested our method on the M$^3$D \cite{m3d}, Twitter-15 \cite{tweet-2015}, Twitter-17 \cite{tweet-2017} and MNRE \cite{mre}. The M$^3$D dataset is multimodal, multilingual, multitask on biographical topics, including two modalities: text and video, two languages: English and Chinese, and four tasks: entity recognition, entity chain extraction, relation extraction, and visual grounding. It contains four entity types and 34 relation types.
Twitter-15 and Twitter-17 are two datasets for MNER, which include user posts and images on Twitter during 2014-2015 and 2016-2017, respectively. They also contains four entity types.
The MNRE dataset is a manually-annotated dataset for MRE task, where the texts and images are crawled from Twitter and a subset of Twitter-15 and Twitter-17. It contains 23 relation types.
Their detailed statistics of the dataset are shown in Table \ref{tab:dataset_Statistics_m3d} and Table \ref{tab:dataset_Statistics_mnre}.

\subsection{Baselines} 
We compared six competitive baseline models, including those using natural language templates and code-style templates. 
\subsubsection{Natural Language Template Models}
\textbf{Qwen3-VL-8B} \cite{qwen3} is a multimodal large language model (MLLM) that can achieve image, text, and video understanding capabilities comparable to flagship models on consumer-grade GPUs. \textbf{REAMO} \cite{multimodal_uie_2} is a framework that can analyze any IE tasks over various modalities, along with their fine-grained groundings. There are also two variants of our method.
\subsubsection{Code-style Templates Models} \textbf{CodeIE} \cite{CodeIE} achieves information extraction by defining corresponding Python function templates for different tasks. \textbf{KnowCoder} \cite{code_uie_1} achieves information extraction by designing a large code-style pattern library containing more than 30,000 knowledge types and a two-stage effective learning framework. \textbf{CodeKGC} \cite{Codekgc} reconstructs natural language text into code format and introduces Chain-of-Thought (CoT), using a code-LLM to extract structured triples from the text. 
\textbf{Code4UIE} \cite{Code4UIE} uses defined classes to structure task representations and dynamically incorporates semantically or structurally similar examples as prompts to guide the large language model in generating code-style output conforming to a specified format.

\begin{table}[!t]
\fontsize{7}{9}\selectfont
\setlength{\tabcolsep}{0.8mm}
\centering
\caption{Comparison of our model with baseline models on M$^3$D. We only report the F1 for each task. \textcolor{black}{\textbf{Bold}} indicates the best result in each column. Avg. represents the average performance of the four tasks.
}
\resizebox{0.45\textwidth}{!}{
\begin{tabular}{c c l c c c c c c c c c c}
\hline
&\phantom{}&&\phantom{}&{\bf Ent.}&\phantom{}&{\bf Cha.}&\phantom{}&{\bf Rel.}&\phantom{}&{\bf Gro.}&\phantom{}&{\bf Avg.}\\
\hline
\multirow{7}{*}{\bf EN}&\phantom{}&Qwen3-VL& \phantom{}&72.07&\phantom{}&67.54&\phantom{}&33.62&\phantom{}&40.13&\phantom{}&53.34\\
&\phantom{}&REAMO & \phantom{}&74.84&\phantom{}&67.06&\phantom{}&34.17&\phantom{}&44.85&\phantom{}&55.23\\
\cdashline{2-13}
&\phantom{}&CodeIE & \phantom{}&77.27&\phantom{}&68.95&\phantom{}&35.99&\phantom{}&46.26&\phantom{}&57.12\\
&\phantom{}&KnowCoder & \phantom{}&78.31&\phantom{}&69.43&\phantom{}&37.72&\phantom{}&46.10&\phantom{}&57.89\\
&\phantom{}&CodeKGC & \phantom{}&77.04&\phantom{}&68.41&\phantom{}&36.51&\phantom{}&45.09&\phantom{}&56.76\\
&\phantom{}&Code4UIE & \phantom{}&77.43&\phantom{}&68.35&\phantom{}&35.70&\phantom{}&45.14&\phantom{}&56.66\\
\cdashline{2-13}
&\phantom{}&Ours&\phantom{}&\bf 79.56&\phantom{}&\textbf{72.85}&\phantom{}&\textbf{41.49}&\phantom{}&\textbf{50.22}&\phantom{}&\textbf{61.03}\\
\hline
\multirow{7}{*}{\bf ZH}&\phantom{}
&Qwen3-VL&\phantom{}&{76.03}&\phantom{}&{75.47}&\phantom{}&{40.22}&\phantom{}&{23.84}&\phantom{}&{53.89}\\
&\phantom{}&REAMO & \phantom{} & 78.62&\phantom{}&76.71&\phantom{}&41.66&\phantom{}&27.64&\phantom{}&56.16\\
\cdashline{2-13}
&\phantom{}&CodeIE & \phantom{}&80.39&\phantom{}&78.06&\phantom{}&43.48&\phantom{}&28.08&\phantom{}&57.50\\
&\phantom{}&KnowCoder & \phantom{}&80.85&\phantom{}&78.45&\phantom{}&44.04&\phantom{}&27.58&\phantom{}&57.73\\
&\phantom{}&CodeKGC & \phantom{}&80.31&\phantom{}&77.76&\phantom{}&43.78&\phantom{}&27.77&\phantom{}&57.41\\
&\phantom{}&Code4UIE & \phantom{}&79.84&\phantom{}&77.10&\phantom{}&42.91&\phantom{}&27.87&\phantom{}&56.93\\
\cdashline{2-13}
&\phantom{}&Ours &\phantom{} & \bf 82.26&\phantom{}&\textbf{80.81}&\phantom{}&\textbf{46.73}&\phantom{}&\textbf{32.14}&\phantom{}&\textbf{60.49}\\
\hline
\end{tabular}
}
\label{tab:main_result}
\end{table}

\subsection{Evaluation Metrics}
We use precision (P), recall (R), and F1 to evaluate our models on all tasks.

In the entity recognition task, if the token sequence and type of a predicted entity are exactly the same as those of a
golden entity, the predicted entity is regarded as true-positive.

In the entity chain extraction task, we follow Tang et al. \cite{chain_Metrics} to use the average of three indicators: MUC, B$^3$ and CEAF to evaluate the prediction entity chain results. 

In the relation extraction task, we evaluate entity relations at the chain level\footnote{In the MNRE dataset, entity relations can be viewed as entity chain relations that contain only one entity.}. When given a predicted relation triplet $r_{pred}$ and a golden relation triplet $r_{gold}$, then when $l_{pred}^{sub} \cap l_{gold}^{sub} \neq \emptyset$, $l_{pred}^{obj} \cap l_{gold}^{obj} \neq \emptyset$ and $t_{pred}^{r} = t_{gold}^{r}$, the predicted relation triplet is regarded as true-positive.

In the visual grounding task, we follow the evaluation method of Yu et al. \cite{introduction_multi_ner}. When the IoU score of the predicted visual area and the golden visual area is greater than 0.5 and the type is the same, then the predicted visual area is regarded as true-positive.

\subsection{Implementation Details}
Our LLM used \texttt{Qwen3-8B} \cite{qwen3} and the visual pre-training model uses \texttt{vit-base} \cite{vit} on all datasets. The training parameters of our model are as follows: the training epoch is 3 for the MNRE dataset and 5 for other datasets, the learning rate is 2e-4 for the English datasets and 8e-4 for Chinese datasets, the batch size is 4 for all datasets. Our model is implemented using PyTorch and trained using an NVIDIA A100-SXM4-80GB GPU. We fine-tuned all baseline models using LoRA \cite{LoRA}. Since the code-style baseline models are all in the pure text domain, we also use the visual pre-trained model \texttt{vit-base} to introduce visual information to them.

Existing code-style baseline models are mainly applicable to entity recognition and relation extraction tasks, but neglect entity chain extraction and visual grounding. Therefore, in our experiments on the M$^3$D dataset, we defined new templates for these two tasks, modeled after the baseline models. 

\begin{table}[!t]
\fontsize{13}{15}\selectfont
\setlength{\tabcolsep}{0.8mm}
\centering
\caption{Comparison of our model with baseline models on Twitter-15, Twitter-17 and MNRE. We only report the F1 for each task.
}
\resizebox{0.4\textwidth}{!}{
\begin{tabular}{l c c c c c c}
\hline
&\phantom{}&\bf Twitter-15&\phantom{}&\bf Twitter-17 &\phantom{} & \bf MNRE\\
\hline
Qwen3-VL-8B&\phantom{}&71.91&\phantom{}&83.27&\phantom{}&66.27\\
REAMO&\phantom{}&72.68&\phantom{}&83.73&\phantom{}&67.14\\
\hdashline
CodeIE&\phantom{}&73.12&\phantom{}&84.93&\phantom{}&69.80\\
KnowCoder&\phantom{}&74.02&\phantom{}&86.47&\phantom{}&70.66\\
CodeKGC&\phantom{}&73.62&\phantom{}&85.25&\phantom{}&69.68\\
Code4UIE&\phantom{}&73.47&\phantom{}&85.36&\phantom{}&69.90\\
\hdashline
Code-MIE (Ours)&\phantom{}&\textcolor{black}{\bf 76.04}&\phantom{}&\textcolor{black}{\bf 88.07}&\phantom{}&\textcolor{black}{\bf 73.94}\\
\hline
\end{tabular}
}
\label{tab:main_result_mner}
\end{table}

\section{Results and Analysis}
\label{Results and Analysis}
\subsection{Main Results}
The results of our method compared with baseline models on three datasets are shown in Table \ref{tab:main_result} and Table \ref{tab:main_result_mner}. 
There are the following observations:
1) Our method achieves state-of-the-art performance on all four datasets. Specifically, on the English and Chinese datasets of M$^3$D, the average F1 score is 61.03\% and 60.49\%, respectively, representing improvements of 3.14\% (61.03\% - 57.89\%) and 2.76\% (60.49\% - 57.73\%) compared to the state-of-the-art baseline model. On the Twitter-15, Twitter-17, and MNRE datasets, the F1 score is 76.04\%, 88.07\%, and 73.94\%, respectively, representing improvements of 2.02\% (76.04\% - 74.02\%), 1.60\% (88.07\% - 86.47\%) and 3.28\% (73.94\% - 70.66\%) compared to the state-of-the-art baseline model, demonstrating the effectiveness of our method. 
2) We found that on all four datasets, the code-style template model outperformed the natural language template model, indicating that code-style templates are better suited for structured tasks such as information extraction than natural language templates. Furthermore, our method achieves at least a 3.36\% performance improvement over the best baseline model for natural language templates.
3) Even among models using the same code-style template, our method remains optimal, perhaps because we added additional knowledge. Furthermore, we found that although the baseline models of the four code-style templates used both Python classes and Python functions to define the templates, the performance difference was not significant. This indicates that the way the templates are defined does not have a major impact on performance.

\begin{table}[!t]
\fontsize{7}{9}\selectfont
\setlength{\tabcolsep}{0.8mm}
\centering
\caption{Ablation experiment on the M$^3$D dataset. (EA: Entity Attribute, SG: Scene Graph, VF: Visual Feature, NLT: Natural Language Template)
}
\resizebox{0.45\textwidth}{!}{
\begin{tabular}{l l c c c c c c c c c c}
\hline
{}&{}&\phantom&{\bf Ent.}&\phantom{}&{\bf Cha.}&\phantom{}&{\bf Rel.}&\phantom{}&{\bf Gro.}&\phantom{}&{\bf Avg.}\\
\hline
\multirow{6}{*}{\bf EN}&Full Model &\phantom{} &\textcolor{black}{\textbf{79.56}}&\phantom{}&\textcolor{black}{\textbf{72.85}}&\phantom{}&\textcolor{black}{\textbf{41.49}}&\phantom{}&\textcolor{black}{\textbf{50.22}}&\phantom{}&\textcolor{black}{\textbf{61.03}}\\
\cdashline{2-12}
&\quad $w/o$ EA &\phantom{}&78.71&\phantom{}&72.55&\phantom{}&39.20&\phantom{}&49.20&\phantom{}&59.92\\

&\quad $w/o$ SG &\phantom{}&78.82&\phantom{}&72.64&\phantom{}&39.69&\phantom{}&48.58&\phantom{}&59.93\\

&\quad $w/o$ VF&\phantom{}&78.56&\phantom{}&71.71&\phantom{}&40.25&\phantom{}&48.91&\phantom{}&59.86\\

&\quad NLT  &\phantom{}&76.28&\phantom{}&72.08&\phantom{}&39.16&\phantom{}&47.38&\phantom{}&58.73\\
\hline
\multirow{6}{*}{\bf ZH}&Full Model &\phantom{} & \textcolor{black}{\textbf{82.26}} &\phantom{}&\textcolor{black}{\textbf{80.81}}&\phantom{}&\textcolor{black}{\textbf{46.73}}&\phantom{}&\textcolor{black}{\textbf{32.14}}&\phantom{}&\textcolor{black}{\textbf{60.49}}\\
\cdashline{2-12}
&\quad $w/o$ EA &\phantom{}&81.47&\phantom{}&80.45&\phantom{}&44.20&\phantom{}&31.27&\phantom{}&59.35\\

&\quad $w/o$ SG &\phantom{}&81.15&\phantom{}&80.47&\phantom{}&43.76&\phantom{}&29.87&\phantom{}&58.81\\

&\quad $w/o$ VF&\phantom{}&80.52&\phantom{}&79.76&\phantom{}&44.55&\phantom{}&30.40&\phantom{}&58.81\\

&\quad NLT &\phantom{}&75.40&\phantom{}&73.08&\phantom{}&43.11&\phantom{}&28.74&\phantom{}&55.08\\
\hline
\end{tabular}
}
\label{tab:abs_m3d}
\end{table}

\begin{table}[!t]
\fontsize{13}{15}\selectfont
\setlength{\tabcolsep}{0.8mm}
\centering
\caption{Ablation on Twitter-15, Twitter-17 and MNRE.
}
\resizebox{0.35\textwidth}{!}{
\begin{tabular}{l c c c c c c}
\hline
&\phantom{}&\bf Twitter-15&\phantom{}&\bf Twitter-17&\phantom{}&\bf MNRE\\
\hline
Full Model &\phantom{} & \textcolor{black}{\bf 76.04}&\phantom{}&\textcolor{black}{\bf 88.07} &\phantom{}&\textcolor{black}{\bf 73.94}\\
\hdashline
\quad $w/o$ EA& \phantom{}&75.53&\phantom{}&87.58&\phantom{}&73.16\\

\quad $w/o$ SG &\phantom{}&75.19&\phantom{}&87.11&\phantom{}&72.28\\

\quad $w/o$ VF&\phantom{}&75.00&\phantom{}&86.94&\phantom{}&71.74\\

\quad NLT &\phantom{}&74.96&\phantom{}&86.92&\phantom{}&73.20\\
\hline
\end{tabular}
}
\label{tab:abs_mner}
\end{table}

\subsection{Ablation Analysis} 
We performed ablation analysis on our model, and the specific results are shown in Table \ref{tab:abs_m3d} and Table \ref{tab:abs_mner}.

1) $w/o$ EA indicates the removal of entity attribute knowledge from our framework. The model's performance degrades across all three datasets, with a significant drop in performance on the relation extraction task of the M$^3$D dataset. This suggests that entity attribute knowledge helps to better identify relations.

2) $w/o$ SG indicates that removing scene graph knowledge from our framework. Similarly, the model's performance degrades across all three datasets, with the most significant drops observed in relation extraction and the basic vision task of the M$^3$D dataset. This suggests that explicit visual objects and their relations contribute to better recognition of textual relations and visual objects.

3) $w/o$ VF indicates that removing visual features from our framework. Compared to attribute and scene graph knowledge, the removal of visual features leads to a greater performance degradation on all three datasets, which is reasonable as it results in the loss of rich multimodal information.

4) NLT indicates that replacing our code-style template with a natural language template results in a significant performance decrease on the M$^{3}$D dataset, with a decrease of 2.30\% (61.03\% - 57.73\%) in English and 5.41\% (60.49\% - 55.08\%) in Chinese. This is because the natural language template has more severe hallucination errors than the code-style template. However, the performance drop was less severe on the Twitter-15, Twitter-17 and MNRE datasets, decreasing by 1.08\% (76.04\% - 74.96\%), 1.15\% (88.07\% - 86.92\%) and 0.74\% (73.94\% - 73.20\%) respectively. This may be because the tasks on these two datasets are simpler, and the constructed templates are shorter, making them less prone to hallucination errors.

\begin{figure}[!t]
\centering
\includegraphics[width=0.45\textwidth]{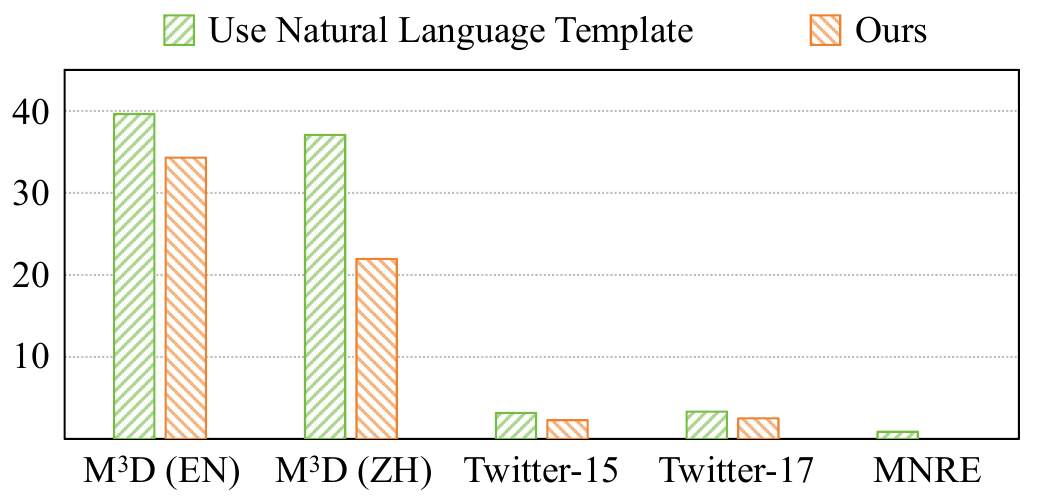}
\caption{
Comparison of the natural language template and code-style template in hallucination errors. 
}
\label{code-style-error}
\end{figure}

\subsection{The Effect of Code-style Template on Hallucination Error Alleviation}
\label{Hallucination Error}
The hallucination errors\footnote{The calculation methods for hallucination error rates are as follows: hallucination error rate = number of samples with hallucination errors / total number of samples} of the two templates on the test set are shown in Figure \ref{code-style-error}. 
We have the following observations:
(1) On all three datasets, models using code-style templates exhibited lower hallucination error rates. This is because code-style templates have stricter grammatical rules, forcing models to follow specific output patterns, while natural language templates lack explicit structural constraints, potentially leading to the generation of explanatory text, the addition of extra modifiers, or incorrect inferences. 
(2) Furthermore, we note that code-style templates resulted in a greater reduction in hallucination errors on the Chinese dataset of M$^3$D compared to the English dataset of M$^3$D. This is primarily due to the lack of alignment between the Chinese and English datasets in M$^3$D. The Chinese dataset contains fewer entities, entity chains, relations, and visual objects, resulting in shorter overall output templates that are less prone to hallucination errors. 
(3) Similarly, on the Twitter-15, Twitter-17 and MNRE datasets, the simpler tasks led to shorter templates, resulting in significantly fewer hallucination errors compared to the M$^3$D dataset.

\begin{figure*}[!t]
\centering
\includegraphics[width=0.95\textwidth]{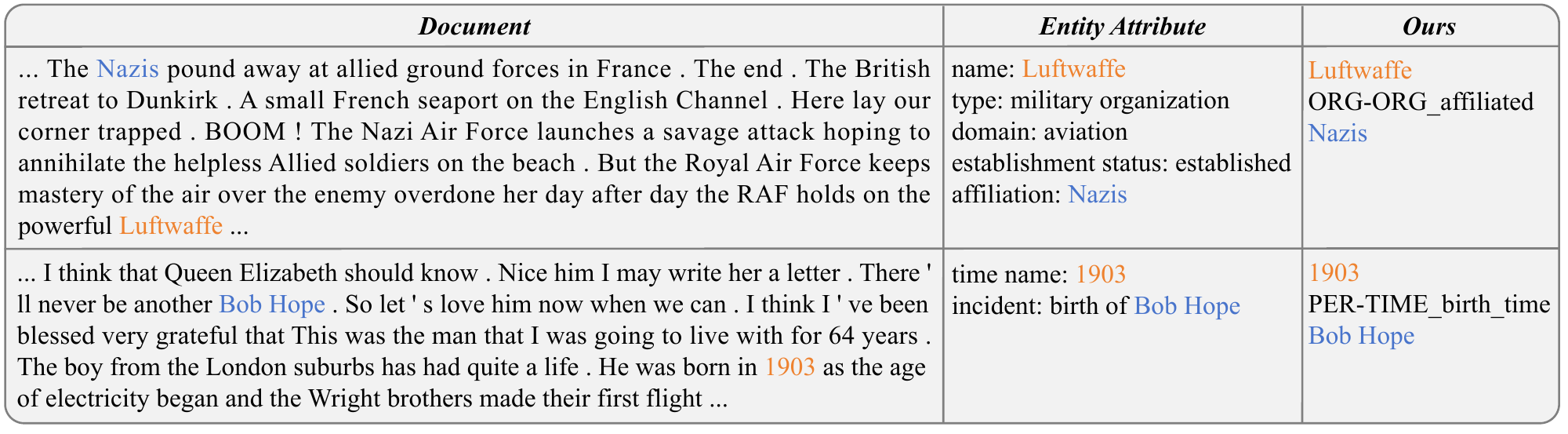}
\caption{Case study of entity attribute knowledge on the M$^3$D dataset.
} 
\label{case_study_ea}
\end{figure*}

\subsection{Case Study in Relation Extraction} 
\label{Case Study}
\subsubsection{Entity Attribute} 

We analyzed the facilitating effect of entity attribute knowledge on the frame on the M$^3$D dataset. Figure \ref{case_study_ea} illustrates two examples: In the first example, the entity attribute knowledge explicitly indicates a "affiliation" relation between "Luftwaffe" and "Nazis," prompting our frame to identify ORG-ORG\_affiliated relation. In the second example, the entity attribute knowledge explicitly indicates that the event "birth of Bob Hope" occurred at the time "1903", prompting our frame to identify PER-TIME\_birth\_time, relationships that were not identified in frames without entity attribute knowledge.

\subsubsection{Scene Graph} 
\begin{figure}[!t]
\centering
\includegraphics[width=0.45\textwidth]{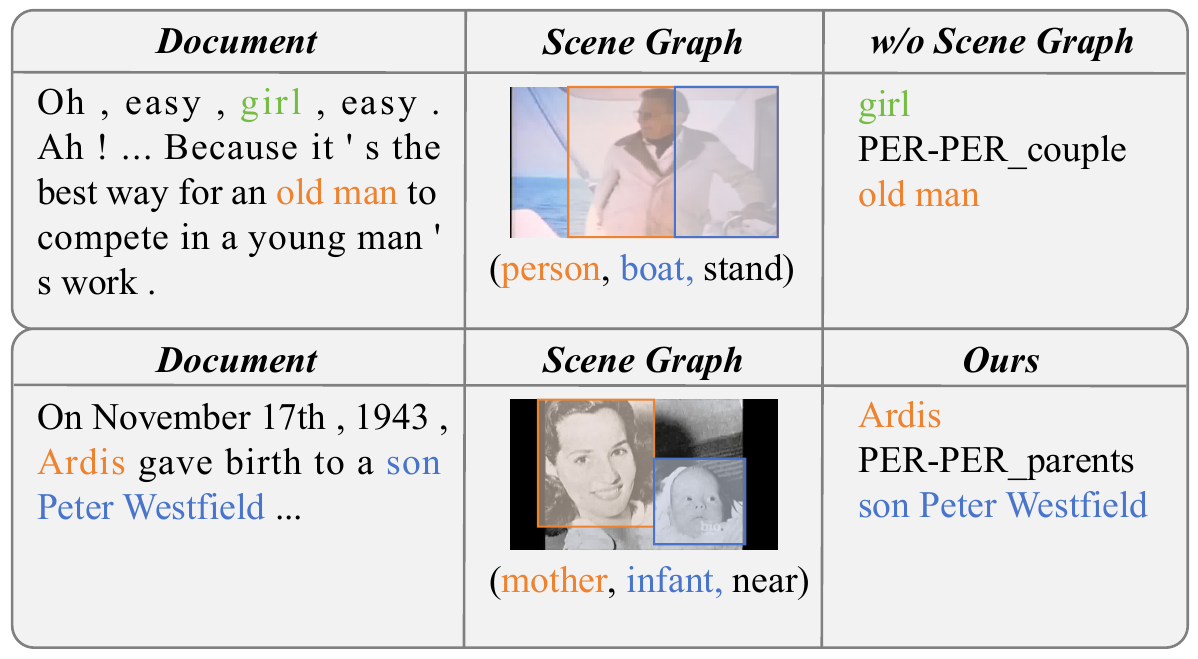}
\caption{Case study of scene graph knowledge  on the M$^3$D dataset.
} 
\label{case_study}
\end{figure}

We analyzed the role of scene graph knowledge in promoting our framework on the M$^3$D dataset. Two examples are shown in Figure \ref{case_study}. For the first example, the scene graph knowledge explicitly represents that "person" and "boat" have a "stand" relation, but not a relation with other person. Therefore, our Code-MIE framework did not recognize any relation, while the ours framework without scene graph knowledge mistakenly recognized that "old man" and "girl" have a PER-PER\_couple relation. In the second example, the scene graph knowledge explicitly represents that "mother" and "infant" have a "near" relation, which means that these two people are likely to have some kind of relation. Therefore, our Code-UNIE framework identified that "Ardis" and "son Peter Westfield" have a PER-PER\_parents relation, while the ours framework without scene graph knowledge does not recognize any relation.

\section{Error Analysis}
\label{Error Analysis}

\begin{figure}[!t]
\centering
\includegraphics[width=0.45\textwidth]{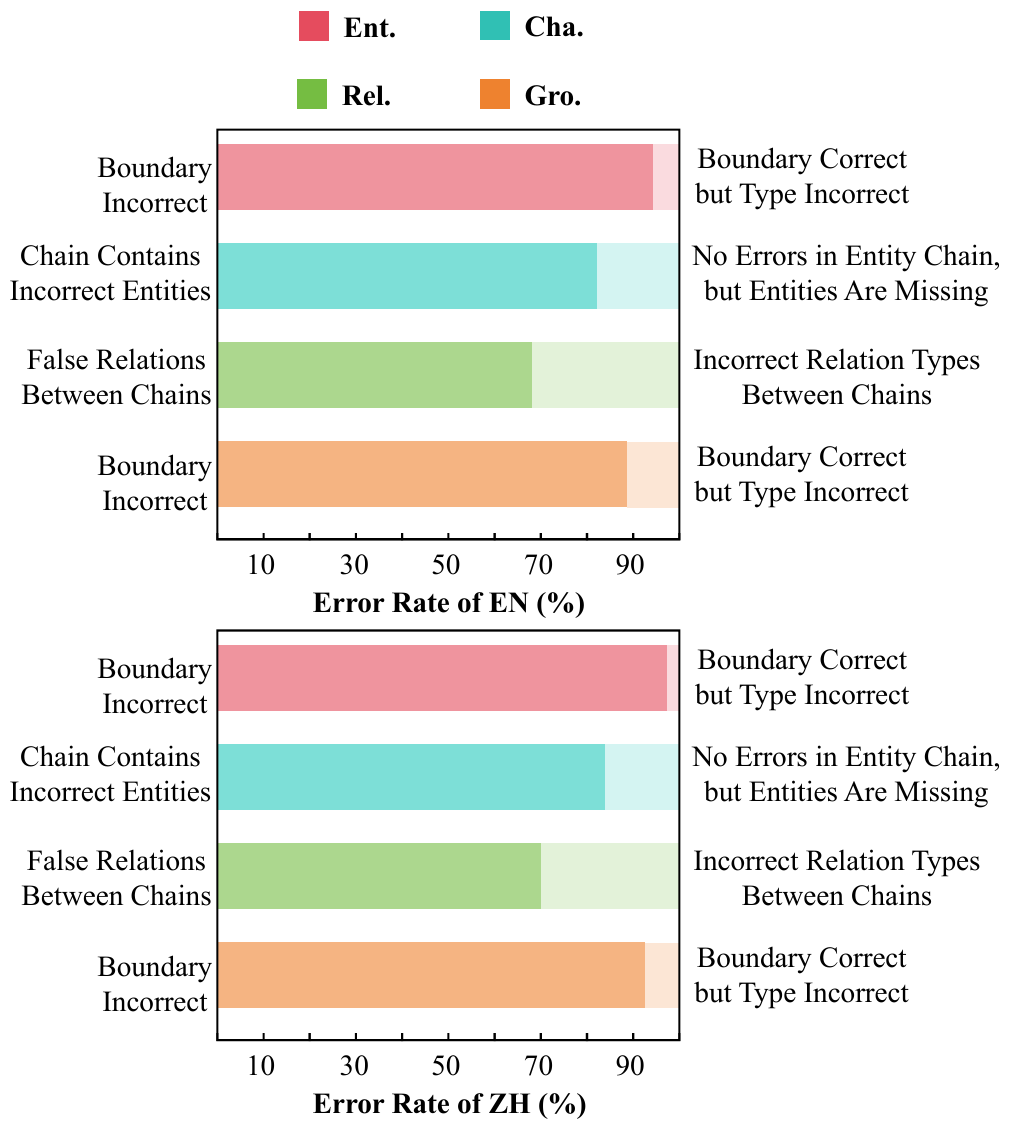}
\caption{Error analysis of four tasks on the M$^3$D dataset.} 
\label{error}
\end{figure}

We performed an error analysis on our model on the M$^3$D dataset. There are two types of errors for each task. The error rate refers to the ratio of the number of each error type to the total number of errors in the prediction for each task. The specific results are shown in Figure \ref{error}.

For the entity recognition task, the errors in model prediction are mainly due to the error in entity boundary recognition. The difficulty in identifying entity boundaries is a well-known problem in previous work \cite{bourd_error_1, bourd_error_2}.

For the entity chain extraction task, we found that the error ratio of the chain contains the incorrect entities is higher, which indicates that our model is more likely to classify entities with similar meanings into one category, resulting in the entity chain contains entities that do not belong to the entity chain.

For the relation extraction task, although the errors in the model prediction are mainly due to the recognition of some entity chain pairs that have no relation, the proportion of relation type prediction errors is also much higher than the minor errors in other tasks. One reason is that relation extraction requires deeper reasoning, and another reason is that there are far more relation types than entity types.

For the visual grounding task, the main reason for the model prediction error is the same as that of the entity recognition task, which is reasonable. The visual grounding task can be regarded as an entity recognition task in vision, so the boundary is still difficult to identify.

\begin{figure}[!t]
\centering
\includegraphics[width=0.5\textwidth]{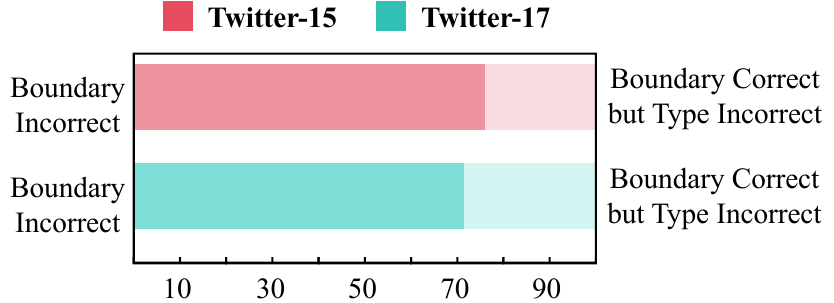}
\caption{Error analysis of four tasks on the Twiter15, Twitter17 and MNRE dataset.} 
\label{error_other}
\end{figure}

We also conducted error analysis on Twitter15, Twitter17 and MNRE. The results are shown in Figure \ref{error_other}. On the Twitter15 and Twitter17 datasets, the main errors are still from boundary incorrect, which is consistent with the results of the entity recognition task on the M$^3$D dataset. On the MNRE dataset, the proportions of the two types of errors are similar, with the proportion of incorrect relation types between entities slightly higher, which is opposite to the results of the relation extraction task on the M$^3$D dataset. This is because the entity pairs on the MNRE dataset are known, and there will not be too many unrelated entity pairs.

\section{Conclusion}
\label{Conclusion}
In this paper, we propose a unified multimodal information extraction framework for code-style, which formalizes the multimodal information extraction task as a Python function. The framework converts images into scene graphs and visual features to explicitly and implicitly contain objects and their relations. We then construct input and output templates based on scene graphs, visual features, and raw text to train an LLM. We conduct experiments on four datasets, and the results show that our framework outperforms all baseline models. We also compare with natural language templates, and the results show that code-style templates are more suitable for structured tasks such as information extraction. In addition, we also conduct a more in-depth analysis of our framework. Finally, we hope that our work can attract more and more research attention in this field.

\section*{Acknowledgments}
This work is supported by the National Key Research and Development Program of China (No. 2022YFB3103602). This work is also funded by Kuaishou.

\bibliography{main}

@inproceedings{multimodal_uie_2,
  author       = {Meishan Zhang and
                  Hao Fei and
                  Bin Wang and
                  Shengqiong Wu and
                  Yixin Cao and
                  Fei Li and
                  Min Zhang},
  title        = {Recognizing Everything from All Modalities at Once: Grounded Multimodal
                  Universal Information Extraction},
  booktitle    = {Findings of the Association for Computational Linguistics: ACL 2024},
  pages        = {14498--14511},
  year         = {2024}
}

@inproceedings{multimodal_ie_3,
  author       = {Philipp Seeberger and
                  Dominik Wagner and
                  Korbinian Riedhammer},
  title        = {{MMUTF:} Multimodal Multimedia Event Argument Extraction with Unified
                  Template Filling},
  booktitle    = {Findings of the Association for Computational Linguistics: EMNLP 2024},
  pages        = {6539--6548},
  year         = {2024},
}

@inproceedings{multimodal_ie_2,
  author       = {Li Yuan and
                  Yi Cai and
                  Jin Wang and
                  Qing Li},
  title        = {Joint Multimodal Entity-Relation Extraction Based on Edge-Enhanced
                  Graph Alignment Network and Word-Pair Relation Tagging},
  booktitle    = {Proceedings of the AAAI Conference on Artificial Intelligence},
  pages        = {11051--11059},
  year         = {2023},
}

@inproceedings{Entity_Attributes_ent_rel,
  title={Prototype-Guided Multimodal Relation Extraction based on Entity Attributes},
  author={Zhang, Zefan and Zhang, Weiqi and Li, Yanhui and Bai, Tian},
  booktitle={Proceedings of the AAAI Conference on Artificial Intelligence},
  volume={39},
  number={24},
  pages={26003--26011},
  year={2025}
}

@inproceedings{multimodal_ie_1,
  title={An Effective Span-based Multimodal Named Entity Recognition with Consistent Cross-Modal Alignment},
  author={Xu, Yongxiu and Xu, Hao and Huang, He-Yan and Cui, Shiyao and Tang, Minghao and Wang, Longzheng and Xu, Hongbo},
  booktitle={Proceedings of the 2024 Joint International Conference on Computational Linguistics, Language Resources and Evaluation (LREC-COLING 2024)},
  pages={1063--1072},
  year={2024}
}

@inproceedings{code_uie_2,
  title={Retrieval-augmented code generation for universal information extraction},
  author={Guo, Yucan and Li, Zixuan and Jin, Xiaolong and Liu, Yantao and Zeng, Yutao and Liu, Wenxuan and Li, Xiang and Yang, Pan and Bai, Long and Guo, Jiafeng and others},
  booktitle={CCF International Conference on Natural Language Processing and Chinese Computing},
  pages={30--42},
  year={2024}
}

@article{m3d,
  title={M3D: a Multimodal, Multilingual and Multitask Dataset for Grounded Document-level Information Extraction},
  author={Liu, Jiang and Li, Bobo and Yang, Xinran and Yang, Na and Fei, Hao and Zhang, Mingyao and Li, Fei and Ji, Donghong},
  journal={IEEE Transactions on Pattern Analysis and Machine Intelligence},
  year={2025},
}

@inproceedings{CodeIE,
  title={CodeIE: Large Code Generation Models are Better Few-Shot Information Extractors},
  author={Li, Peng and Sun, Tianxiang and Tang, Qiong and Yan, Hang and Wu, Yuanbin and Huang, Xuan-Jing and Qiu, Xipeng},
  booktitle={Proceedings of the 61st Annual Meeting of the Association for Computational Linguistics (Volume 1: Long Papers)},
  pages={15339--15353},
  year={2023}
}

@inproceedings{Knowcoder-x,
  title={Knowcoder-x: Boosting multilingual information extraction via code},
  author={Zuo, Yuxin and Jiang, Wenxuan and Liu, Wenxuan and Li, Zixuan and Bai, Long and Wang, Hanbin and Zeng, Yutao and Jin, Xiaolong and Guo, Jiafeng and Cheng, Xueqi},
  booktitle={Findings of the Association for Computational Linguistics: ACL 2025},
  pages={14486--14509},
  year={2025}
}

@inproceedings{retrieval_Classification,
  title={Retrieval over Classification: Integrating Relation Semantics for Multimodal Relation Extraction},
  author={Hei, Lei and Liao, Tingjing and Qi, Yiyang and Wang, Jiaqi and Li, Ruiting and Ren, Feiliang and others},
  booktitle={Proceedings of the 2025 Conference on Empirical Methods in Natural Language Processing},
  pages={18689--18704},
  year={2025}
}

@inproceedings{code_uie_1,
  author       = {Zixuan Li and
                  Yutao Zeng and
                  Yuxin Zuo and
                  Weicheng Ren and
                  Wenxuan Liu and
                  Miao Su and
                  Yucan Guo and
                  Yantao Liu and
                  Lixiang Lixiang and
                  Zhilei Hu and
                  Long Bai and
                  Wei Li and
                  Yidan Liu and
                  Pan Yang and
                  Xiaolong Jin and
                  Jiafeng Guo and
                  Xueqi Cheng},
  title        = {KnowCoder: Coding Structured Knowledge into LLMs for Universal Information
                  Extraction},
  booktitle    = {Proceedings of the 62nd Annual Meeting of the Association for Computational
                  Linguistics (Volume 1: Long Papers)},
  pages        = {8758--8779},
  year         = {2024},
}

@inproceedings{uie_more_study_2,
  title={RUIE: Retrieval-based Unified Information Extraction using Large Language Model},
  author={Liao, Xincheng and Duan, Junwen and Huang, Yixi and Wang, Jianxin},
  booktitle={Proceedings of the 31st International Conference on Computational Linguistics},
  pages={9640--9655},
  year={2025}
}

@inproceedings{uie_more_study_1,
  author={Qi, Yunjia and Peng, Hao and Wang, Xiaozhi and Xu, Bin and Hou, Lei and Li, Juanzi},
  title        = {{ADELIE:} Aligning Large Language Models on Information Extraction},
  booktitle    = {Proceedings of the 2024 Conference on Empirical Methods in Natural
                  Language Processing},
  pages        = {7371--7387},
  year={2024}
}

@inproceedings{chain_Metrics,
  title={Parallel Data Helps Neural Entity Coreference Resolution},
  author={Tang, Gongbo and Hardmeier, Christian},
  booktitle={Findings of the Association for Computational Linguistics: ACL 2023},
  pages={3162--3171},
  year={2023}
}

@inproceedings{lora,
  title={LoRA: Low-Rank Adaptation of Large Language Models},
  author={Hu, Edward J and Wallis, Phillip and Allen-Zhu, Zeyuan and Li, Yuanzhi and Wang, Shean and Wang, Lu and Chen, Weizhu and others},
  booktitle={International Conference on Learning Representations},
  pages={1--13},
  year={2021}
}

@inproceedings{more,
  title={Named Entity and Relation Extraction with Multi-Modal Retrieval},
  author={Wang, Xinyu and Cai, Jiong and Jiang, Yong and Xie, Pengjun and Tu, Kewei and Lu, Wei},
  booktitle={Findings of the Association for Computational Linguistics: EMNLP 2022},
  pages={5925--5936},
  year={2022}
}

@inproceedings{Mixup_Image_Augmentation_mner,
  title={Enhancing Multimodal Named Entity Recognition through Adaptive Mixup Image Augmentation},
  author={Xu, Bo and Jiang, Haiqi and Wei, Jie and Jing, Hongyu and Du, Ming and Song, Hui and Wang, Hongya and Xiao, Yanghua},
  booktitle={Proceedings of the 31st International Conference on Computational Linguistics},
  pages={1802--1812},
  year={2025}
}

@inproceedings{Hierarchical_mner,
  title={Hierarchical aligned multimodal learning for NER on tweet posts},
  author={Liu, Peipei and Li, Hong and Ren, Yimo and Liu, Jie and Si, Shuaizong and Zhu, Hongsong and Sun, Limin},
  booktitle={Proceedings of the AAAI Conference on Artificial Intelligence},
  volume={38},
  number={17},
  pages={18680--18688},
  year={2024}
}

@article{vit,
  title={Visual transformers: Token-based image representation and processing for computer vision},
  author={Wu, Bichen and Xu, Chenfeng and Dai, Xiaoliang and Wan, Alvin and Zhang, Peizhao and Yan, Zhicheng and Tomizuka, Masayoshi and Gonzalez, Joseph and Keutzer, Kurt and Vajda, Peter},
  journal={arXiv preprint arXiv:2006.03677},
  year={2020}
}

@inproceedings{mrc_mner,
  title={Query prior matters: A MRC framework for multimodal named entity recognition},
  author={Jia, Meihuizi and Shen, Xin and Shen, Lei and Pang, Jinhui and Liao, Lejian and Song, Yang and Chen, Meng and He, Xiaodong},
  booktitle={Proceedings of the 30th ACM international conference on multimedia},
  pages={3549--3558},
  year={2022}
}

@inproceedings{bourd_error_1,
  title={Rethinking boundaries: End-to-end recognition of discontinuous mentions with pointer networks},
  author={Fei, Hao and Ji, Donghong and Li, Bobo and Liu, Yijiang and Ren, Yafeng and Li, Fei},
  booktitle={Proceedings of the AAAI conference on artificial intelligence},
  volume={35},
  number={14},
  pages={12785--12793},
  year={2021}
}

@article{bourd_error_2,
  title={TOE: A grid-tagging discontinuous NER model enhanced by embedding tag/word relations and more fine-grained tags},
  author={Liu, Jiang and Ji, Donghong and Li, Jingye and Xie, Dongdong and Teng, Chong and Zhao, Liang and Li, Fei},
  journal={IEEE/ACM Transactions on Audio, Speech, and Language Processing},
  volume={31},
  pages={177--187},
  year={2022},
  publisher={IEEE}
}

@inproceedings{Scene_Graph,
  title={FACTUAL: A Benchmark for Faithful and Consistent Textual Scene Graph Parsing},
  author={Li, Zhuang and Chai, Yuyang and Zhuo, Terry Yue and Qu, Lizhen and Haffari, Gholamreza and Li, Fei and Ji, Donghong and Tran, Quan Hung},
  booktitle={Findings of the Association for Computational Linguistics: ACL 2023},
  pages={6377--6390},
  year={2023}
}

@inproceedings{introduction_multi_ner,
  title={Grounded multimodal named entity recognition on social media},
  author={Yu, Jianfei and Li, Ziyan and Wang, Jieming and Xia, Rui},
  booktitle={Proceedings of the 61st Annual Meeting of the Association for Computational Linguistics (Volume 1: Long Papers)},
  pages={9141--9154},
  year={2023}
}

@inproceedings{related_work_ner_entity_level,
  title={Entity-level interaction via heterogeneous graph for multimodal named entity recognition},
  author={Zhao, Gang and Dong, Guanting and Shi, Yidong and Yan, Haolong and Xu, Weiran and Li, Si},
  booktitle={Findings of the Association for Computational Linguistics: EMNLP 2022},
  pages={6345--6350},
  year={2022}
}

@inproceedings{related_work_ner_1,
  title={Improving Multimodal Named Entity Recognition via Entity Span Detection with Unified Multimodal Transformer},
  author={Yu, Jianfei and Jiang, Jing and Yang, Li and Xia, Rui},
  booktitle={Proceedings of the 58th Annual Meeting of the Association for Computational Linguistics},
  pages={3342--3352},
  year={2020}
}

@inproceedings{tweet-2015,
  title={Visual attention model for name tagging in multimodal social media},
  author={Lu, Di and Neves, Leonardo and Carvalho, Vitor and Zhang, Ning and Ji, Heng},
  booktitle={Proceedings of the 56th Annual Meeting of the Association for Computational Linguistics (Volume 1: Long Papers)},
  pages={1990--1999},
  year={2018}
}

@inproceedings{tweet-2017,
  title={Adaptive co-attention network for named entity recognition in tweets},
  author={Zhang, Qi and Fu, Jinlan and Liu, Xiaoyu and Huang, Xuanjing},
  booktitle={Proceedings of the AAAI Conference on Artificial Intelligence},
  pages={5674--5681},
  year={2018}
}

@inproceedings{mre,
  title={Mnre: A challenge multimodal dataset for neural relation extraction with visual evidence in social media posts},
  author={Zheng, Changmeng and Wu, Zhiwei and Feng, Junhao and Fu, Ze and Cai, Yi},
  booktitle={2021 IEEE International Conference on Multimedia and Expo (ICME)},
  pages={1--6},
  year={2021}
}

@inproceedings{related_Information,
  title={Information Screening whilst Exploiting! Multimodal Relation Extraction with Feature Denoising and Multimodal Topic Modeling},
  author={Wu, Shengqiong and Fei, Hao and Cao, Yixin and Bing, Lidong and Chua, Tat-Seng},
  booktitle={Proceedings of the 61st Annual Meeting of the Association for Computational Linguistics (Volume 1: Long Papers)},
  pages={14734--14751},
  year={2023}
}

@inproceedings{related_Translation,
  title={Rethinking Multimodal Entity and Relation Extraction from a Translation Point of View},
  author={Zheng, Changmeng and Feng, Junhao and Cai, Yi and Wei, Xiaoyong and Li, Qing},
  booktitle={Proceedings of the 61st Annual Meeting of the Association for Computational Linguistics (Volume 1: Long Papers)},
  pages={6810--6824},
  year={2023}
}

@article{qwen3,
  title={Qwen3 technical report},
  author={Yang, An and Li, Anfeng and Yang, Baosong and Zhang, Beichen and Hui, Binyuan and Zheng, Bo and Yu, Bowen and Gao, Chang and Huang, Chengen and Lv, Chenxu and others},
  journal={arXiv preprint arXiv:2505.09388},
  year={2025}
}

@inproceedings{Code4UIE,
  title={Retrieval-augmented code generation for universal information extraction},
  author={Guo, Yucan and Li, Zixuan and Jin, Xiaolong and Liu, Yantao and Zeng, Yutao and Liu, Wenxuan and Li, Xiang and Yang, Pan and Bai, Long and Guo, Jiafeng and others},
  booktitle={CCF International Conference on Natural Language Processing and Chinese Computing},
  pages={30--42},
  year={2024},
}

@article{Codekgc,
  title={Codekgc: Code language model for generative knowledge graph construction},
  author={Bi, Zhen and Chen, Jing and Jiang, Yinuo and Xiong, Feiyu and Guo, Wei and Chen, Huajun and Zhang, Ningyu},
  journal={ACM Transactions on Asian and Low-Resource Language Information Processing},
  volume={23},
  number={3},
  pages={1--16},
  year={2024},
}
\bibliographystyle{IEEEtran}

\end{document}